\documentclass[runningheads]{llncs}
\usepackage{todonotes}
\usepackage{graphicx}
\usepackage{siunitx}
\begin{document}
\title{Playing Soccer without Colors in the SPL: \\A Convolutional Neural Network Approach}
\titlerunning{Playing Soccer without Colors in the SPL}

\author{Francisco Leiva\inst{\star} \and Nicol\'as Cruz\thanks{These authors contributed equally to this work}  \and Ignacio Bugue\~no \and Javier Ruiz-del-Solar}
\institute{Advanced Mining Technology Center \& Dept. of Elect. 
Eng., Universidad de Chile
\email{\{francisco.leiva, nicolas.cruz, ignacio.bugueno, jruizd\}@ing.uchile.cl}
}
\maketitle  
\begin{abstract}
The goal of this paper is to propose a vision system for humanoid robotic soccer that does not use any color information. The main features of this system are: (i) real-time operation in the NAO robot, and (ii) the ability to detect the ball, the robots, their orientations, the lines and key field features robustly. Our ball detector, robot detector, and robot's orientation detector obtain the highest reported detection rates. The proposed vision system is tested in a SPL field with several NAO robots under realistic and highly demanding conditions. The obtained results are: robot detection rate of 94.90\%, ball detection rate of 97.10\%, and a completely perceived orientation rate of 99.88\% when the observed robot is static, and 95.52\% when the observed robot is moving.

\end{abstract}
\section{Introduction}
The perception of the environment is one of the key abilities for playing soccer; without an adequate vision system it is not possible to determine the position of field's features or to self-localize. It is also impossible to determine the position of the ball and the other players, which is necessary in order to play properly. Given that the soccer environment is highly dynamic and has a predefined physical setup, most of the current vision systems use color information.

In the case of the SPL and the former Four-Legged League, the first generation of vision systems analyzed colored objects which were then segmented. Year by year, the restriction of having colored objects in the field was relaxed: (i) the number of colored beacons was first reduced and then beacons were not used anymore, (ii) the goals were first colored and solid, then non-solid, and finally white, (iii) the ball used to be orange, and since 2016, black and white. However, still most of the teams use color information to detect field features (e.g., lines and their intersections), other players and the ball. Very recently, Convolutional Neural Networks (CNNs) have been used for detecting the robots and/or the ball (e.g., \cite{CruzTsunekawa,RobotDetection2016,BallHeat2016,StoneBall}), but even in these cases, the CNN-based detectors require object proposals which are usually obtained using color information. Therefore, to the best of our knowledge, color-free vision systems have not been used in robotic soccer, at least not in the SPL. Some of the main reasons are: (i) the challenge of achieving real-time operation when using limited computational resources, (ii) the problem of training deep detectors without having very large databases, and (iii) the challenge of having fast and color-free object proposals.

The goal of this paper is to propose a color-free vision system for the SPL. The main features of this system are: (i) real-time operation in the NAO robot, and (ii) the ability to detect the ball, the robots, their orientations, the lines and key field features very robustly. In fact, our ball, robots and robots' orientation detectors are highly performant; they obtain the highest reported detection rates.

\section{Playing Soccer without Color Information}
\label{sec:visionSystem}

In this section we present the proposed vision system. Section \ref{sec:generalFramework} broadly explains the general characteristics and functioning of the vision framework, while Sections \ref{sec:highContrastDetection}, \ref{sec:robotDetection}, \ref{sec:robotOrientation}, \ref{sec:ballDetection} and \ref{sec:ffeaturesDetection} detail the operation of each of its main modules.

\subsection{The General Framework}
\label{sec:generalFramework}

The main feature of our framework is that it manages to detect the ball, other players, their orientations, and key features of the field without using any color information: all the processing is performed on grayscale images. This is done by following a cascade methodology (inspired in \cite{CascadeMethodology}) that combines classical approaches widely used in pattern recognition and modern CNN-based classifiers.

\begin{figure}
\includegraphics[width=\textwidth]{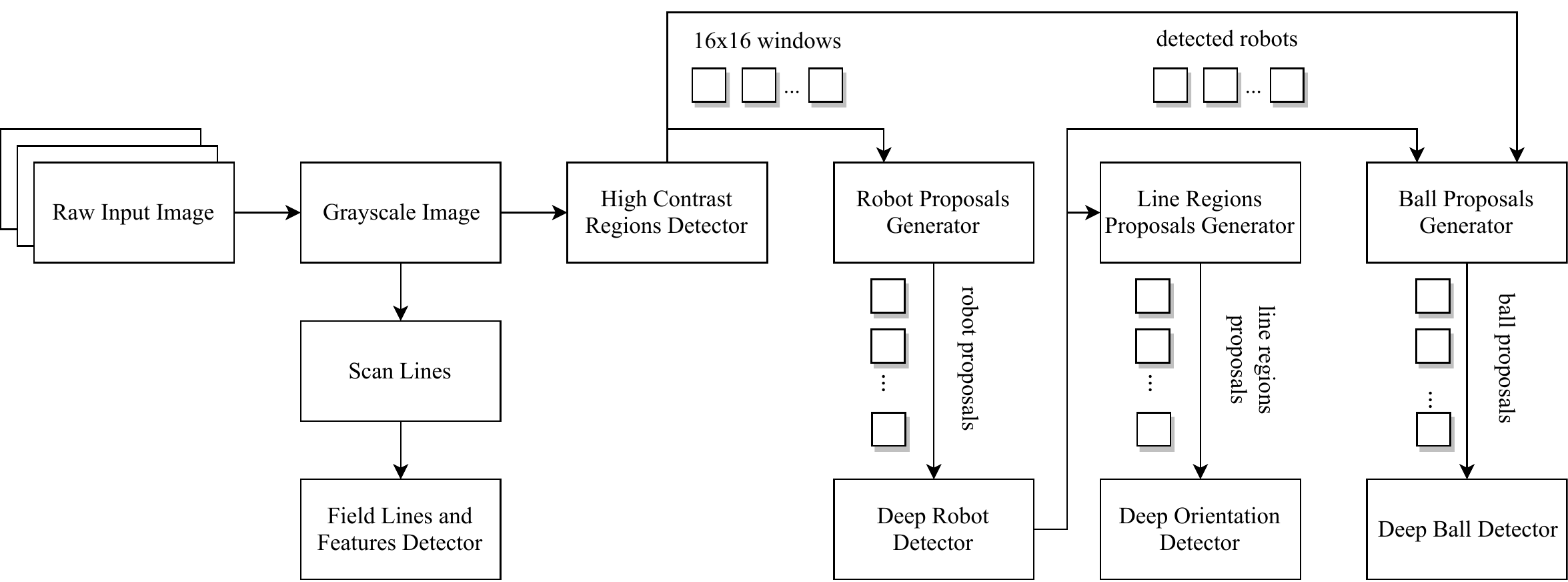}
\caption{Block diagram of the proposed vision system.} 
\label{fig:visionFramework}
\end{figure}

The proposed vision framework is illustrated in Fig. \ref{fig:visionFramework}. While the detection of lines and field features is done by using a set of rules and heuristics, both the detection of the ball and the other robots is done by means of object proposals and their subsequent classification using CNNs. This cascade approach takes advantage of the information previously extracted from the image to use it in benefit of following processing modules. 

\subsection{High Contrast Regions Detection}
\label{sec:highContrastDetection}

Since the robots and the ball used in the SPL possess high contrast, an effective approach to know where to search for them is to find high contrast regions in the images. To do this, the grayscale input images are scanned using windows of 16$\times$16 pixels. Regions outside the field boundaries and within the body of the observer robot are discarded. The remaining windows are used to construct histograms of pixels, which are used to estimate thresholds for image binarization using Otsu's method \cite{Otsu}. Windows with thresholds over a predefined value are considered as important, since they may be close or within another robot or the ball. Since the chosen threshold for the selection of windows could be restrictive and leave out image regions belonging to objects of interest, a morphological dilation operation is applied on the previously selected windows, which means that all the 16x16 pixels blocks adjacent to selected windows are also considered as high contrast regions.

\subsection{Robot Detection}
\label{sec:robotDetection}

In \cite{CruzTsunekawa} we presented a robot detector based on CNNs, capable of operating in real time. The system was based on the classification of color-based robot proposals (generated by B-Human's robot perceptor \cite{bhumancoderelease}). This was modeled as a binary classification problem where proposals could be labeled as robots or non-robots. The system processed hypotheses in $\sim$1 ms with an average accuracy of $\sim$97\%. 
Although this system achieved a very high performance, it possessed some major drawbacks. First, while the CNN classifier was very robust to noise and variations of the illumination, the same did not apply to the color-based robot proposal generator. Adverse environmental conditions could lead the algorithm to produce an excessive amount of object hypotheses, or none at all. The second drawback derived from the CNN inference time of $\sim$1 ms. While such a network is deployable on a NAO robot, it is much slower than alternative algorithms based on heuristics or shallow classifiers, and can be prohibitively slow when too many robot proposals are generated. In this paper we address both problems by changing the robot proposals generation approach, and by further reducing the inference times while maintaining the detection accuracy.

The proposal generation of this new framework does not use any color information: it uses vertical scan lines over all the image $x$-coordinates where high contrast regions were detected (see Section \ref{sec:highContrastDetection}). The scan lines search for luminance changes in order to find the robots' feet positions, and by performing geometric sanity checks, the proposal generator provides a set of bounding boxes which may contain the robots' body. Most checks are similar to the rules used in the B-Human player detector \cite{bhumancoderelease}, but applied to a grayscale image. This approach is more robust to changes in lighting since it relies on contrast information rather than heuristic color segmentation.

The obtained grayscale image regions are then fed to a CNN, which we call \textit{RobotNet}, that classifies the proposals as robots or non-robots. This CNN is based on the architecture described in Section \ref{sec:squeeze}. Using grayscale image regions allows the system to perform in real time for a large number of robot proposals, since the reduction of input channels greatly reduces the CNN's inference time.

\subsection{Robot Orientation Determination}
\label{sec:robotOrientation}
\begin{figure}
\includegraphics[width=\textwidth]{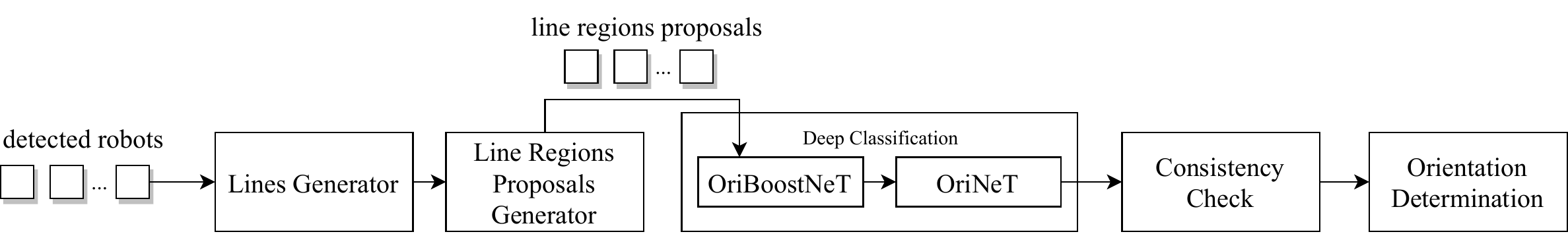}
\caption{Robot orientation module pipeline.} \label{fig:rotationPipeline}
\end{figure}

Inspired on \cite{BHumanOrientation}, we propose an improved Vision-Based Orientation Detection for the SPL League, which makes use of CNNs in order to achieve much better prediction accuracy than the original system. The general architecture of the module is presented in Fig. \ref{fig:rotationPipeline}.

This system uses the bounding boxes of the \textit{Detected Robots} as inputs. Over these regions, the set of points that compose the robots' lower silhouette \cite{BHumanOrientation} is calculated by the \textit{Lines Generator} module, which extracts a region corresponding to the robot's feet and analyses its Contrast-Normalized Sobel (CNS) image \cite{CNSImageRef} by using vertical scan lines. Over each scan line pixel an horizontal median filter is applied and its response is compared to a threshold. Pixels with a filter response below the threshold are considered as part of the lower silhouette. Then, by iterating for each scan line, the subset of points that make up a closed convex region can be obtained by using Andrew's convex hulls algorithm \cite{AndrewAlgorithm}. For each consecutive pair of points of the convex set we calculate a line model in field coordinates. Each line model is then validated with the set of points of the lower silhouette, by using a voting methodology akin to the RANSAC algorithm \cite{RANSAC}. The line with the higher number of votes is selected as the \textit{first line}. Once the linear model has been chosen, a \textit{second line} may be generated by iterating over the remaining pairs of convex points. This line must comply with a series of conditions such as a minimum and maximum length and approximate orthogonality to the first line in order to be accepted as valid.

To estimate the orientation of the observed robot, the lines are classified to determine the robot's direction. To do this, a region that includes the robot's feet and legs is constructed around each line by the \textit{Line Regions Proposals Generator} module. The regions are then classified by the \textit{Deep Classification} module which is based on CNNs, whose structure is shown in Fig. \ref{fig:firestuff}. For each of the line's regions a CNN that measures its quality, \textit{OriBoostNet}, is first applied. Regions with too much motion blur or that were incorrectly estimated are discarded to decrease the number of wrong orientation estimations. If a region is accepted, it is then fed to a second CNN, \textit{OriNet}, that in turns classifies it as a side, front or back region. Afterwards, we perform a \textit{Consistency Check} by imposing that no more than one region of each class must exist. This further reduces the number of incorrect orientation estimations. Finally, the \textit{Orientation Determination} is performed by combining the rotation given by the inverse tangent from two points belonging to the analyzed line, with the direction of the line determined by its class. The resulting orientation is added to a buffer that stores the last 11 measurements and a circular median filter is applied over it. In order to avoid invalid results, we consider the direction as valid only for a small period of time if no new samples are added to the buffer.

\subsection{Ball Detection}
\label{sec:ballDetection}

\begin{figure}
\centering
\includegraphics[width=\textwidth]{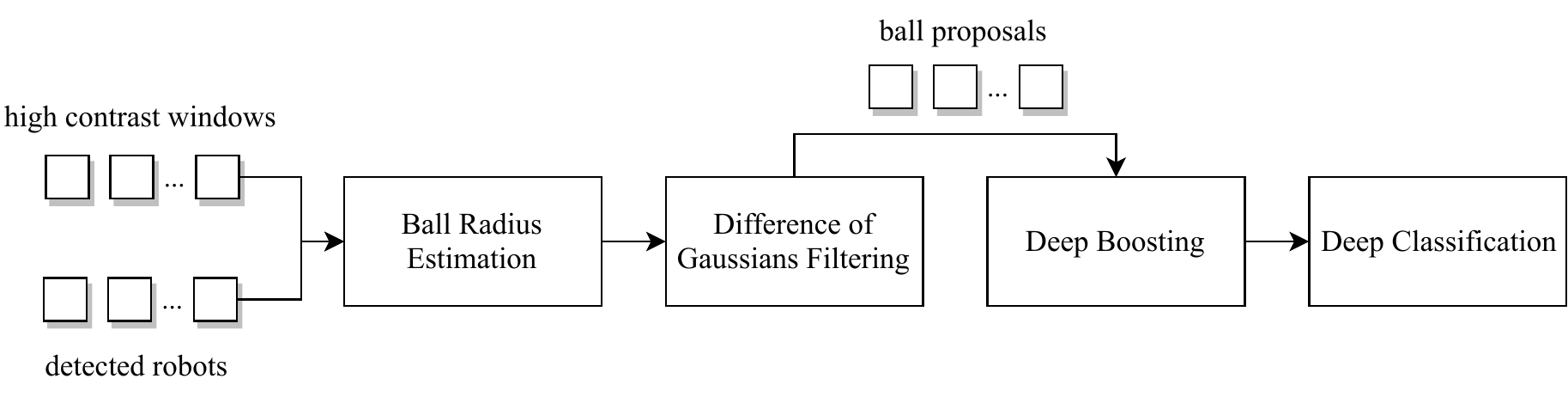}
\caption{Ball detection module pipeline.} \label{fig:ballPipeline}
\end{figure}

In the proposed vision framework, the ball detector follows the paradigm of proposal generation and subsequent classification. Fig. \ref{fig:ballPipeline} shows the general architecture of this module.

Our ball proposal generator is inspired on the hypothesis provider developed by the HTWK team \cite{htwkrr}. The main differences between both approaches are: (i) we only use grayscale images, (ii) we use a different method to estimate high contrast regions (see Section \ref{sec:highContrastDetection}), and (iii) we use the robots' detections in order to improve the generation of proposals.

The proposal generator uses the high contrast regions and the robots' detections to provide the ball hypotheses. To accomplish this task, the generator performs a pixel-wise scan over all image windows that were detected by the high contrast detector and over image regions corresponding to the detected robots' feet. During this stage, a \textit{Ball Radius Estimation} is calculated for every analyzed position in image coordinates. 

The next stage consists in a \textit{Difference of Gaussians} (DoG)\textit{ Filtering}. During this process, DoG filters' local responses are calculated for each scan coordinate. The support regions of the filters are dependent on the estimated ball radii, so we are actually searching for blobs by means of the same approach used by the SIFT algorithm \cite{sift}. Additional DoG responses are calculated in front of the other robots' feet given that the ball may be in these regions. Finally, only the highest responses are used to construct a set of proposals, whose size depends on the estimation of the radius of the ball.

To perform the ball detection, the proposals are fed to a cascade of two CNNs which classifies them as ball or non-ball. The first CNN, \textit{BoostBallNet}, performs \textit{Deep Boosting} to both limit the proposals' number to a maximum of five, and sort them based on their confidence. The second CNN, \textit{BallNet}, performs \textit{Deep Classification}, meaning that it processes the filtered hypotheses to detect the ball. Both networks are extremely fast and accurate, having execution times of 0.043 ms and 0.343 ms, and accuracy rates of 0.965 and 0.984, respectively.

\subsection{Field Lines \& Special Features Detection}
\label{sec:ffeaturesDetection}

The field lines and features detection follow the same algorithm released by B-Human \cite{bhumancoderelease}. The main difference with respect to the original approach, is that in the proposed framework no color information is used. To do this, a set of vertical and horizontal scan lines are used, which save transitions from high-to-low and low-to-high luminance. This allows the detection of a set of points which are then fed to the B-Human's algorithm in order to associate them with lines and other features such as the middle circle, corners and intersections.

\section{Design and Training of the CNN-based Detectors}
\label{sec:designAndTraining}

In this section we focus on the design and training methodologies used to obtain highly performant CNN based classifiers for our vision framework. Section \ref{sec:squeeze} presents the network architectures of our classifiers and Section \ref{sec:trainingmethod} describes the active learning-based algorithm that was developed to train them.

\subsection{Base CNN}
\label{sec:squeeze}

\begin{figure}
\centering
\includegraphics[width=\textwidth]{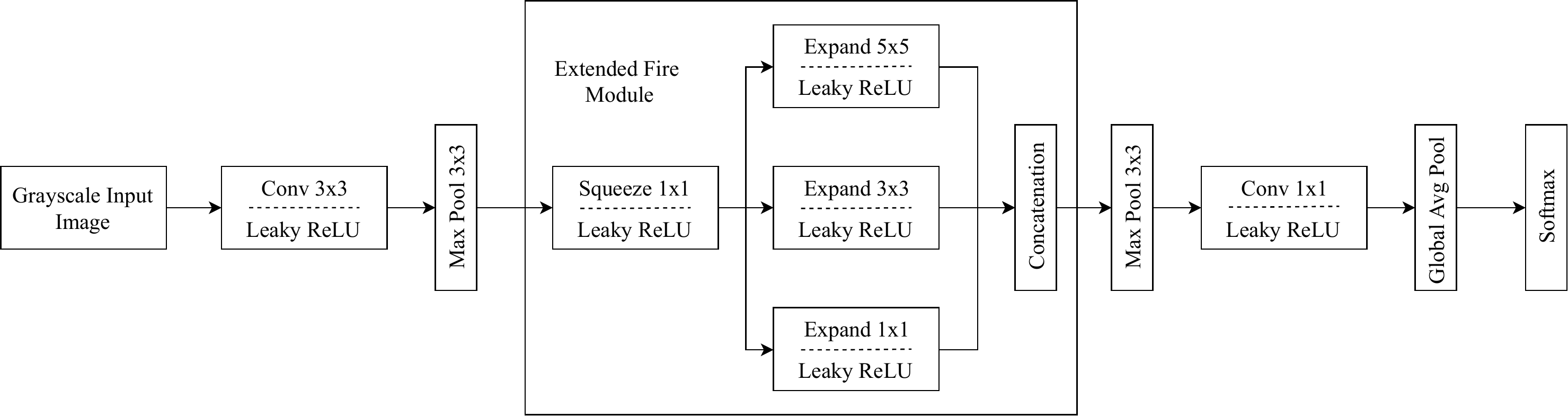}
\caption{Modified MiniSqueezeNet network structure.} \label{fig:firestuff}
\end{figure}

The proposed vision system makes use of several classifiers based on CNNs. While these CNNs are used for different purposes, their architectures remain similar across all the developed modules and are based on the work presented in \cite{CruzTsunekawa}, with slight variations to achieve higher speeds while maintaining accuracy. The main component of these architectures is the extended Fire module, which was developed in \cite{CruzTsunekawa} inspired on the original Fire module proposed in \cite{Squeezenet}. This module concatenates the outputs produced by filters of different sizes in order to achieve increased accuracy while being computationally inexpensive. Small filters are used to extract local information across channels, while bigger filters obtain global information which is more spatially spread out. The information obtained at different scales is then combined into a single tensor and fed to the next layer. This allows the network to extract and work with both local and detailed features as well as broad, global features. Following this approach allows the training of performant models, but concatenating the information of several filters could be prohibitively expensive in terms of computational cost. To account for this, a 1$\times$1 filter is placed at the beginning of each Fire module to compress the size of the representation that correspond to the input of the subsequent larger and more expensive filters. In contrast with the previous miniSqueezeNet version, all newly developed CNNs have grayscale image inputs. Since most of the computational cost of the network correspond to the first convolutional layers, this translates in sharply reduced inference times, and an accuracy loss of about 0.01. Another change is the use of leaky ReLU \cite{Leaky} instead of ReLU as activation functions. Previously, we used ReLU in most layers, however, this sometimes resulted in the ``dying ReLU'' problem while training (no
gradients flow backward through the neurons). The use of leaky ReLU solves this, while incurring in no accuracy loss. All CNNs were developed using the Darknet library \cite{darknet}. A diagram of the new CNN structure is presented in Fig. \ref{fig:firestuff}.

\subsection{Active Learning Training Methodology}
\label{sec:trainingmethod}
In order to train the classifiers, we implemented an active learning-based algorithm that automatically selects and pseudo-annotates unlabeled data. 

We start by initializing the parameters of the CNNs by training them using publicly available datasets (e.g., SPQR datasets \cite{RobotDetection2016}). However, if we directly use the obtained CNN weights in our vision framework, the classifiers behave poorly because there is a distribution mismatch between the samples present in the public datasets, and the ones that our proposal generators output. 

To address this problem, the classifiers must be trained using the same kind of samples that would actually reach the networks during games. To accomplish this, the vision system is deployed on the NAO robot and data is collected using the proposal generators. Each proposal is then stored in the robot's memory with a label annotated by the CNN. To get uncorrelated data, we set a constraint for the object hypotheses to be saved: for the robot proposals, data is acquired periodically in accordance to a predefined time step; for the ball proposals, samples can only be saved if no other proposals with the same position and estimated radius were previously collected. The next stage consists in actively checking the data saved by the observer robot, and manually annotate the samples that were incorrectly labeled. We then aggregate this data to the original dataset and re-train the models. 

The above process is repeated until the CNNs reach a high performance. By doing this, we are progressively aggregating correctly labeled samples to provide enough training data for robust feature learning, but also aggregating hard examples which the models fail to correctly infer, to actively shape the decision boundary of the classifiers.

After we obtain proficient models by following the described methodology, we further enhance them by switching to a bootstrap procedure. To do this, we add confidence-based constrains to collect new training data in environments where the object we want to detect is absent. For instance, if we are getting false positives from the ball detector, we would set the NAO robot to collect data from proposals with high confidence in environments were no balls are present. The samples collected would then be used to re-train the ball classifiers.

This active learning-bootstrap procedure results in a dramatical improvement in the performance of the classifiers after only a few iterations, and also allows the fine tuning of the CNN parameters by means of using data aggregation when an abrupt domain change occurs. Since the inputs to our models have relatively low dimensionality, the space used in the NAO memory during the data collection process is very small, for instance, 1,000 robot proposal samples weight about 3 MB. This procedure, combined with the automatic selection and labeling of the new samples, make the training process extremely time-wise efficient.

\section{Results}
\label{sec:results}

\subsection{CNN Classification}
\label{sec:classifiersPerformance} 
Table \ref{tab:classifiersPerformance} shows the model complexity (number of CNN parameters), average inference time (on the NAO robot), and accuracy for each developed CNN.

Results show that the classifiers achieve very high performance while maintaining low inference times, which proves that their use is suitable for real time applications such as playing soccer. This also validates the effectiveness of the proposed methodology for the design and training of the classifiers. Finally, this also proves that the use of color information is not necessary to detect robots or balls when using expressive classifiers such as CNNs. In fact, the CNN used in the robot detector achieves a similar accuracy rate that the model proposed in \cite{CruzTsunekawa}, while being approximately 2.75 times faster. 

\begin{table}[h]
\centering
\caption{Performance of the developed CNNs.}
\label{tab:classifiersPerformance}
\begin{tabular}{lccccc}
\hline
\textbf{Model}               &  \textbf{RobotNet} & \textbf{BoostBallNet} & \textbf{BallNet} & \textbf{OriBoostNet} & \textbf{OriNet} \\ \hline
Input size         &   24$\times$24$\times$1        &   12$\times$12$\times$1            &   26$\times$26$\times$1       &   12$\times$12$\times$1           & 24$\times$24$\times$1      \\
N° of parameters    &  884        &     125         &   444      &      246       &     657   \\
Inference time (ms) &   0.382       &       0.043       &   0.343      &        0.059     &     0.329   \\
Accuracy            &   0.969       &       0.965       &   0.984      &    0.962         &     0.984   \\ \hline
\end{tabular}
\end{table}

\subsection{Robots, Ball and Field Features Detection Systems}
\label{sec:detectionresults}
For the robots and ball detectors, results are divided on proposal generation and module performance. We replicated typical and challenging game conditions in order to acquire about 600 processed frames by an observer robot. Several lighting conditions were imposed while collecting these frames in order to test the robustness and reliability of our modules. The analysis of these frames allowed the extraction of empirical results in relation to the performance of the proposals generators and each detector, which are shown in Table \ref{tab:detectionPerformance}.

Results show that the robots and ball proposals generators achieve high recall rates, while producing an average number of hypotheses per frame that can be processed in real time by the subsequent classifiers. 
Given the recall rate of the ball proposals module and the percentage of true positives of the boosting stage, the overall detection module has a very high detection rate. In fact, our ball detector outperforms B-Human's implementation, which achieves an overall accuracy rate of 0.697 when testing it under the same conditions.

Finally, the field lines and features detector was tested by comparing the difference between the real and the estimated robot pose. The estimation was obtained by using the field lines and features detected by our module. By using this approach we calculated a mean squared error of 40.07 mm, which indicates that our detector is very accurate and reliable.

\begin{table}
\centering
\caption{Performance of the robots and ball detection systems.}
\label{tab:detectionPerformance}
\begin{tabular}{lcc}
\hline
\textbf{Module}          & \textbf{Robot Detector}$\quad$ & \textbf{Ball Detector} \\ \hline
Avg. Proposals per Frame$\quad$ &         3.05                &     10.3                   \\
Proposals Recall         &          0.972               &             0.993           \\
Overall Accuracy         &            0.949             &              0.971          \\ \hline
\end{tabular}
\end{table}
\vspace{-0.5cm}

\subsection{Robot Orientation Determination}
\label{sec:orientationresults}
\begin{figure}[t]
\includegraphics[trim={1.1cm 1.2cm 1.1cm 0.0cm},clip,width=\textwidth]{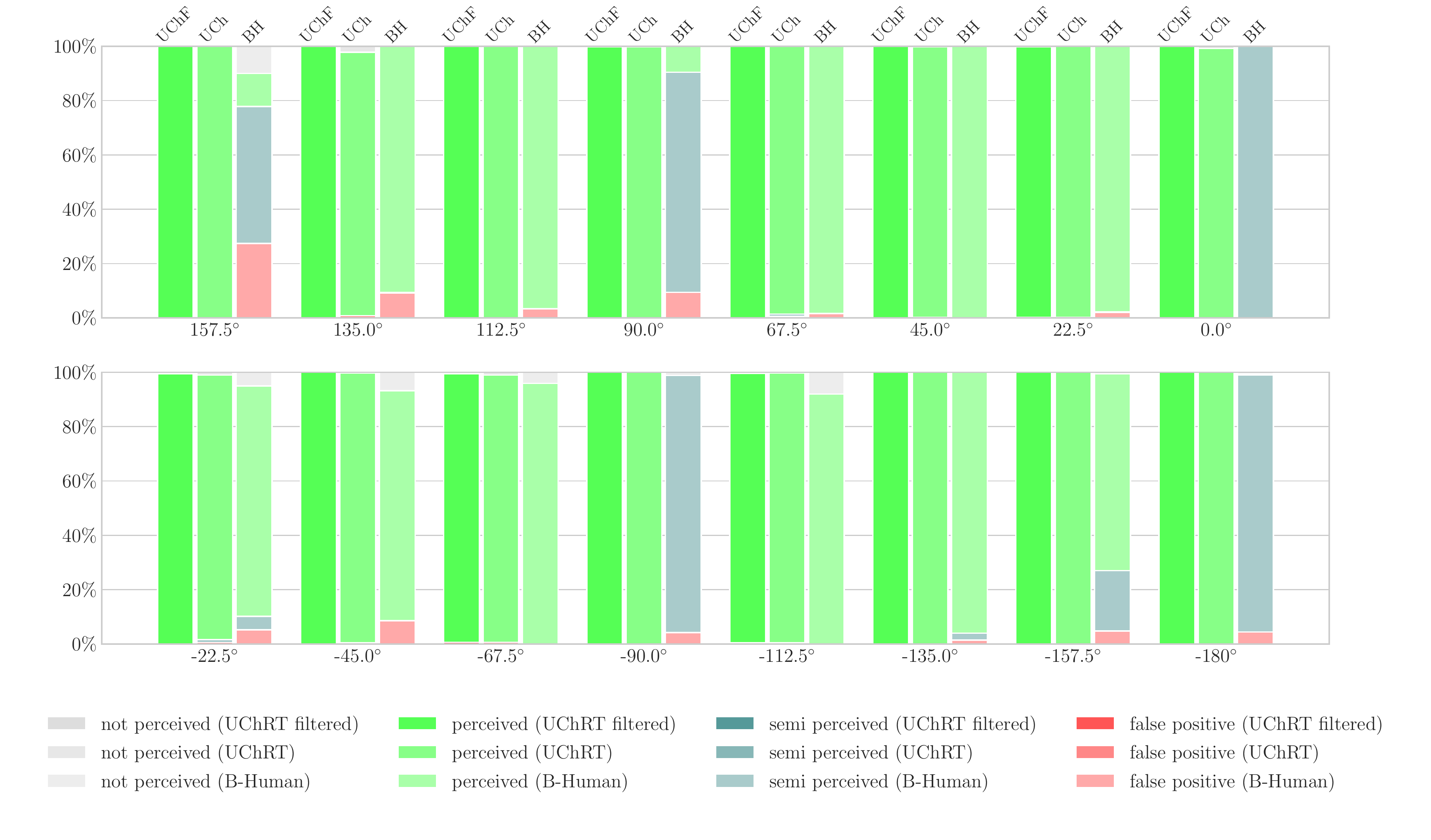}
\caption{Results obtained for the first experiment. Graph shows a performance comparison between raw (UCh) and filtered (UChF) estimations for our orientation detector and a B-Human system replication (BH).}
\label{fig:staticRot}
\end{figure}

In Fig. \ref{fig:staticRot} we present a comparison between the B-Human algorithm proposed in \cite{BHumanOrientation}, our base orientation determination system, and its output after applying a circular median filtering. For this experiment, the observer and the observed robot are static and placed at a distance of 120cm from each other. For each measurement the observed robot was rotated \ang{22.5} around its axis. As in \cite{BHumanOrientation}, we define a \textit{false positive} as any estimation that deviates more than a tolerance angle of \ang{11.25} from the ground-truth. The orientation is classified as \textit{semi perceived} when the rotation can be determined but the facing direction is unknown. The class \textit{not perceived} corresponds to any frame where the orientation could not be calculated, while an orientation estimation is \textit{perceived} if it does not deviate more than a tolerance angle of \ang{11.25} from the ground-truth orientation.

\begin{figure}[h]
\includegraphics[trim={1.0cm 1.2cm 1.0cm 0.1cm},clip,width=\textwidth]{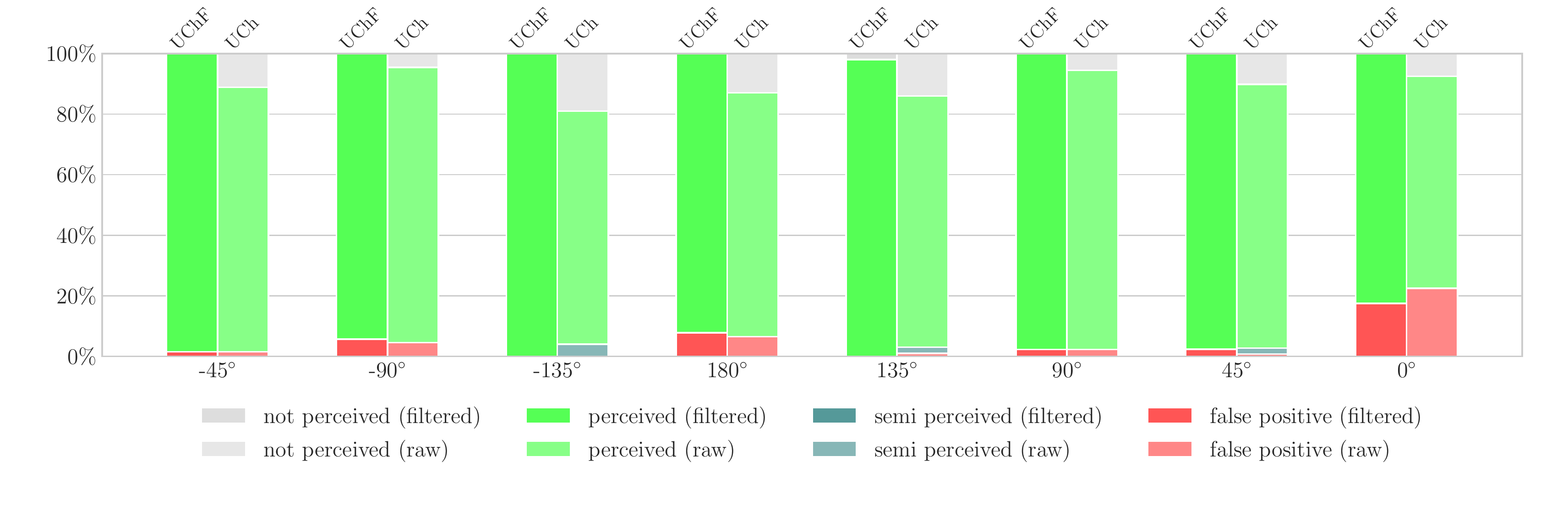}
\caption{Dynamic experiment results. Graph shows a performance comparison between raw (UCh) and filtered (UChF) estimations for our orientation detector.}
\label{fig:comparisonorientation2}
\end{figure}

In Fig. \ref{fig:comparisonorientation2} we show the results obtained when testing our system in a dynamic environment, where the observed robot is moving at a speed of 12.0 cm/s, while the observer remains static. The observed robot is rotated in \ang{45} around its axis for each measurement. We define the same classes for the orientation estimations as in the static experiment, but using a tolerance angle of \ang{22.5}.

As shown in Fig. \ref{fig:staticRot} and Fig. \ref{fig:comparisonorientation2}, the proposed method outperforms the original B-Human implementation. The orientation estimation is completely perceived 99.88\% of the time in static conditions, and 95.52\% of the time in the dynamic experiment. It is clear that the algorithm proposed is better at determining the facing direction of the observed robots. This  results in an increased number of completely perceived orientations while sharply decreasing the number of semi perceived orientations. Also, noise filtering techniques such as the median filter and RANSAC algorithm, combined with the utilization of a CNN contribute to lowering the number of false positive estimations. Finally, the integration of the circular median filter further reduces the number of false positives. 

\subsection{Profiling}
\label{sec:profiling}
Table \ref{tab:profiling} shows the maximum and average execution times for the different modules of the proposed vision framework when deployed on the NAO v5 platform. The results obtained show that the proposed color-free vision system is deployable on platforms with limited processing capacity (such as the NAO robot). In addition, they prove the importance of the dimensionality reduction of CNN-based classifier inputs, and how this design decision impacts the performance of the system from a time-efficiency point of view.

\begin{table}[htpb]
\centering
\caption{Vision framework profiling.}
\label{tab:profiling}
\begin{tabular}{lcc}
\hline
\textbf{Module}                         & \textbf{Max. (ms)}$\quad$ & \textbf{Avg. (ms)} \\ \hline
High Contrast Regions Detector $\quad$&       2.755    &    1.478       \\ 
Field Lines \& Features Detector          &      2.909    &     1.300        \\ 
Robot Proposals Generator      &     2.692      &       1.083    \\ 
Robot Detector                 &       2.417    &        0.939   \\ 
Robot Orientation Detector     &    4.537    &      1.366     \\ 
Ball Proposals Generator       &     2.506      &        1.132   \\ 
Ball Detector                  &     6.959      &      2.452     \\ \hline
\end{tabular}
\end{table}
\vspace{-0.4cm}

\section{Conclusions}
\label{sec:conclusions}

This paper presents a new vision framework that does not use any color information. This is a novel approach for vision systems designed for the SPL, achieving very high performance while being computationally inexpensive.

The proposed vision system we present introduces four new modules: a redesigned robot detector, a visual robot orientation estimator, a brand new ball detector, and finally, a color-free field lines and features detector. All modules developed for this paper are able to run simultaneously in real-time when deployed on the NAO robot, and greatly outperform our previous perception system.

Furthermore, we demonstrate that CNN-based classifiers are a useful tool to solve most of the perception requirements of the SPL, and generally translate in an overall better performance of the corresponding module when coupled with good region proposal algorithms, and proper design and training techniques.

\section*{Acknowledgements}
This work was partially funded by FONDECYT Project 1161500.

\end{document}